\newcommand{\AuthorA}{Boutin}%
\newcommand{\FirstNameA}{Victor}%
\newcommand{\AuthorB}{Franciosini}%
\newcommand{\FirstNameB}{Angelo}%
\newcommand{\AuthorC}{Ruffier}%
\newcommand{\FirstNameC}{Franck}%
\newcommand{\AuthorD}{Perrinet}%
\newcommand{\FirstNameD}{Laurent U}%
\newcommand{\Institute}{Aix Marseille Univ, CNRS, INT, Inst Neurosci Timone, Marseille, France}
\newcommand{\InstituteC}{Aix Marseille Univ, CNRS, ISM, Marseille, France}%
\newcommand{\EmailD}{laurent.perrinet@univ-amu.fr}%
\newcommand{\Title}{%
From biological vision to unsupervised hierarchical sparse coding
}%
\newcommand{\Abstract}{ 
The formation of connections between neural cells is emerging essentially from an unsupervised learning process. For instance, during the development of the primary visual cortex of mammals (V1), we observe the emergence of cells selective to localized and oriented features. This leads to the development of a rough contour-based representation of the retinal image in area V1. \textcolor{black}{We propose a biological model of the formation of this representation along the thalamo-cortical pathway. To} \textcolor{black}{achieve }\textcolor{black}{this goal, we replicated the \textit{Multi-Layer Convolutional Sparse Coding} (ML-CSC) algorithm developed by Michael Elad's group.} 
This type of algorithm alternates (i) a coding phase to encode the information and (ii) a learning phase to find the proper encoder (also called dictionary). 
We trained our network on a database containing images of faces. \textcolor{black}{The receptive fields of the modeled neurons show similarities with their biological counterpart found in V1 and beyond.}

}
\newcommand{\Acknowledgments}{%
This research received funding from the European Union's H2020 programme under the Marie Sk\l{}odowska-Curie grant agreement n\textsuperscript{o}713750 and by the Regional Council of Provence-Alpes-C\^ote d'Azur, A*MIDEX (n\textsuperscript{o}ANR-11-IDEX-0001-02). 
}
\newcommand{\Links}{%
Corresponding author: \EmailD . 
Code and supplementary material available at https://github.com/VictorBoutin/MLCSC. 
} %
\documentclass[a4paper,10pt,twocolumn]{article}

\usepackage[top=1.5cm, left=1.5cm, right=1.5cm, bottom=1.5cm]{geometry}

\renewenvironment{abstract}{\bf\small {\em\ Abstract---}}{}
\usepackage[english]{babel}
\usepackage[utf8]{inputenc}
\usepackage[T1]{fontenc}

\usepackage{amsfonts,amssymb,amsmath,amsthm}
\usepackage{tikz}
\usepackage[footnotesize]{caption}
\usepackage{color}

\usepackage[ruled,vlined]{algorithm2e}
\usepackage[numbers,comma,sort&compress,round]{natbib} %
\usepackage{csquotes}
\usepackage{mathrsfs}
\usepackage{amsfonts}
\usepackage{stmaryrd}
\usepackage{amssymb}
\def\mQuad{\hskip.61\fontdimen6\font}

\usepackage{amsfonts,amssymb,amsmath,amsthm}

\title{\Title}
\author{\FirstNameA\ \AuthorA$^1$, \FirstNameB\ \AuthorB$^1$,  \FirstNameC\ \AuthorC$^2$ ~and \FirstNameD\ \AuthorD$^1$\footnote{\Links}
.\\
  \footnotesize $^1$ \Institute\ , \footnotesize $^2$ \InstituteC\ 
  } \date{\empty} 
\begin{document}
\maketitle
\begin{abstract} 
\Abstract
\end{abstract}
\section{Introduction}
\label{sec:introduction}
Finding an accurate representation to describe concisely a signal (images, sounds, or information) is one of the major concerns of modern machine learning. One of the most successful paradigm to achieve such a representation relies on algorithms performing alternately sparse coding and dictionary learning~\cite{olshausen1996emergence,olshausen1997sparse}. When they are combined with the ability to learn multiple levels of descriptive features in a hierarchical structure, these algorithms can represent more \textcolor{black}{complex} and diverse signals. Such hierarchies have shown great success in tasks such as classification~\cite{bo2011hierarchical} or image compression~\cite{huang2011learning}. In addition, if the structure of the learned dictionaries is convolutional (i.e. each dictionary could be represented by constructing a Toeplitz-structured matrix), then such algorithms better model local patterns that appear anywhere in the signal (e.g. an image) without adding redundancy to the representation~\cite{kavukcuoglu2010learning}.\\
\textcolor{black}{Interestingly, these algorithms developed for machine learning and signal processing are compatible with the organization of the visual cortex. In particular, the \textit{Multi-Layer Convolutional Sparse Coding} (ML-CSC) model, defined by~\cite{Sulam2017} could be reinterpreted in the light of neuroscience.}
Given a set of convolutional dictionaries $\{ D_{i}\}_{i=1}^{L}$ where $D_{i}$ models the synaptic weights between neurons of the ($i-1$)-th and $i$-th layer ($L \geq 2$ is the number of layers of the model) of appropriate dimension, a signal $y \in \mathbb{R}^{N}$ admits a representation in terms of the ML-CSC model if: 
\begin{equation}
\label{eq:eq1}
    \left\{
       \begin{array}{ll}
       y = \gamma_{1} \circledast D_{1}, & \Vert \gamma_{1} \Vert_{0} \leq \lambda_{1} \\
       \gamma_{i-1} = \gamma_{i} \circledast D_{i}, & \Vert \gamma_{i} \Vert_{0} \leq \lambda_{i} \quad \forall i \in \llbracket 2~;~ L \rrbracket 
       \end{array}
	\right.
\end{equation}
\textcolor{black}{where $\circledast$ is the discrete convolution operator,  $\gamma_{i}$ is the sparse representation of the input $\gamma_{i-1}$ in the new basis $D_{i}$, and could be considered as the neuronal response of the layer $i$. The parameter $\lambda_{i}$ measures the number of active coefficients and forces the level of sparsity at level $i$.   } 

In particular, this formulation allows to define an effective dictionary $D^{(i)}$ at the $i^{th}$ layer: 
\begin{equation} 
\label{eq:eq2}
 D^{(i)} = D_{1}\circledast D_{2} \circledast ... \circledast D_{i} 
\end{equation} 
where $D^{(i)}$ could be interpreted as \textcolor{black}{ the set of receptive fields in the input space of the neurons in the $i^{th}$ layer.}  

Within this framework,~\cite{Sulam2017} gives theoretical guarantees of stability and recovery for the learning and coding problem. Given a set of $K$ input signals $\{y_{k} \}_{k=1}^{K}\in \mathbb{R}^{N}$, this problem consists in finding for each $k$ a set of sparse maps $\{ \gamma_{i}^{k} \}_{i=1}^{L}$ and dictionaries $ \mathcal{D} = \{ D_{i} \}_{i=1}^{L}$ that fit the following formulation:
\begin{equation}
\label{eq:eq3}
\left\{
	\begin{array}{lll}
    \displaystyle \min_{\{\gamma^k_{L}\}, \mathcal{D}} \sum_{k=1}^{K} \Vert y_{k} - D^{(L)} \circledast \gamma_{L}^{k}\Vert_{2}^{2} + \sum_{i=2}^{L} \zeta_{i} \Vert D_{i}\Vert_{1} + \lambda_{L} \Vert \gamma_{L}^{k}\Vert_{1} \\
       \displaystyle \text{s.t.} \mQuad \forall i,j \mQuad \Vert d_{i}^{j}\Vert_{2} = 1, \mQuad   \text{with} \mQuad 
  D_{i} = [d_{i}^{1}, d_{i}^{2}, ... , d_{i}^{J}] \\
  	
  \end{array}
	\right.
\end{equation}
where $d_{i}^{j}$ is the $j^{th}$ atom of the $i^{th}$ dictionary. Eq.~(\ref{eq:eq3}) shows that only the deepest layer representation $\gamma_{L}$ is calculated \textcolor{black}{ and sparsity-constrained by the $\lambda_L$ parameter. The constraint on the sparsity of the intermediate-level dictionaries (tuned with the scalars $\zeta_{i}$) prevents intermediate representations from being dense}. If needed, these representations could be inferred quickly in a descending order using: 
\begin{equation}
\label{eq:eq4}
\gamma_{i-1} =  D_{i} \circledast \gamma_{i} 
\end{equation}
\textcolor{black}{The FISTA~\cite{beck2009fast} algorithm is used to determine  $\gamma_{L}$. The \textit{Multi-Layer Convolutional Dictionary Learning } (ML-CDL) was presented in~\cite{Sulam2017} to solve an alternative of eq.~(\ref{eq:eq3}). It includes a weight decay and was applied for classifying MNIST images}.
\textcolor{black}{In this paper, we first detail our implementation of the ML-CDL algorithm that is solving the problem presented in eq.~(\ref{eq:eq3}). We then describe our 2-layered network, that was used on natural images. 
} 
\begin{algorithm}[!b]
\DontPrintSemicolon
	\KwIn{training set  $ \{ y_{k} \}_{k=1}^{K}$, initial dictionaries $ \{ D_{i} \}_{i=1}^{L}$}
 	\For{$k = 1$ \KwTo $K$} {
        
		$D^{(L)} = D_{1} \circledast D_{2} \circledast ... \circledast D_{L-1}$ \;
        $\hat{D}^{(L)} \leftarrow  D^{(L)}/ \text{Norm}(D^{(L)})$\;
        $\gamma_{L}^{k} = \text{SparseCoding}(y_{k}, \hat{D}^{(L)}, \lambda_{L}$)\;
        $\gamma_{L}^{k} \leftarrow \gamma_{L}^{k} / \text{Norm}(D^{(L)})$\;
 		\For{$i =  L$ \KwTo 2} { 	
 			$D_{i} \leftarrow  \mathcal{S}_{\zeta_{i}} (D_{i} - \eta \frac{\partial (\Vert y^{k} - D^{(L)}  \circledast \gamma_{L}^{k}\Vert_{2}^{2})}{\partial D_{i}})$ \;
 			$D_{i} \leftarrow D_{i} / \text{Norm}(D_{i})$ \;
 		}
        $D_{1} \leftarrow D_{1} - \eta \frac{\partial (\Vert y^{k} - D^{(L)}  \circledast \gamma_{L}^{k}\Vert_{2}^{2})}{\partial D_{1}}$\;
 		$D_{1} \leftarrow  D_{1} / \text{Norm}(D_{1})$ \;			
 	}
    \KwOut{$\{ \gamma_{L}^{k} \}_{k=1}^{K},  \{ D_{i} \}_{i=1}^{L}$}
 	
 \caption{ML-CDL}\label{NMLCSC}
\end{algorithm}
\section{ML-CDL: unsupervised hierarchical sparse coding and learning}

We implemented algorithm~\ref{NMLCSC} in Python to solve eq.~(\ref{eq:eq3}) under the assumption of the ML-CSC model. 
First, we compute the effective dictionary of the deepest layer $D^{(L)}$ by using eq.~(\ref{eq:eq2}). This dictionary is used as an input of a sparse coding algorithm (FISTA) to compute the deepest sparse representation $\gamma_{L}$. 
We then have to compensate the non-unitary norm of the effective dictionary $D^{(L)}$ by dividing the sparse map $\gamma_{L}$ by the norm of $D^{(L)}$. The Norm($\cdot$) function outputs the $\ell_{2}$-norm for each atom. The next step consists in finding the set of dictionary  $ \{ D_{i} \}_{i=1}^{L}$ that is minimizing eq.~(\ref{eq:eq3}). This is done by computing the gradient of the $\ell_{2}$-norm term in eq.~(\ref{eq:eq3}) with respect to each dictionary $D_{i}$. \textcolor{black}{The learning rate of the gradient descent is noted $\eta$. }
A soft thresholding operator $\mathcal{S}_{\zeta_{i}}(\cdot)$ is applied to force a certain sparsity level to the deep set of dictionaries. 
We then $\ell_{2}$-normalized the dictionaries atom-wise at each training step.
\section{Experimental Results}
\label{sec:second-section}
We used the AT\&T face database~\cite{ATTDB} made of $400$ grayscale images of $40$ individuals. The faces are centered in each image. We split the data set into $20$ batches composed of $20$ images each; each image was resized to $64\times 64$ pixels. In order to reduce the dependencies between images in the dataset, we performed a Local Contrast Normalization pre-processing \textcolor{black}{by removing the mean of a neighborhood from a particular pixel and dividing it by the variation of the pixel, as described in~\cite{jarrett2009best}}. The result of this preprocessing is shown in Fig.~\ref{fig:fig2}a. 

\begin{figure}[!hb]
	\begin{tikzpicture}
		\centering
		\draw [anchor=north west] (.0\linewidth, .382\linewidth) node {\includegraphics[width=.469\linewidth]{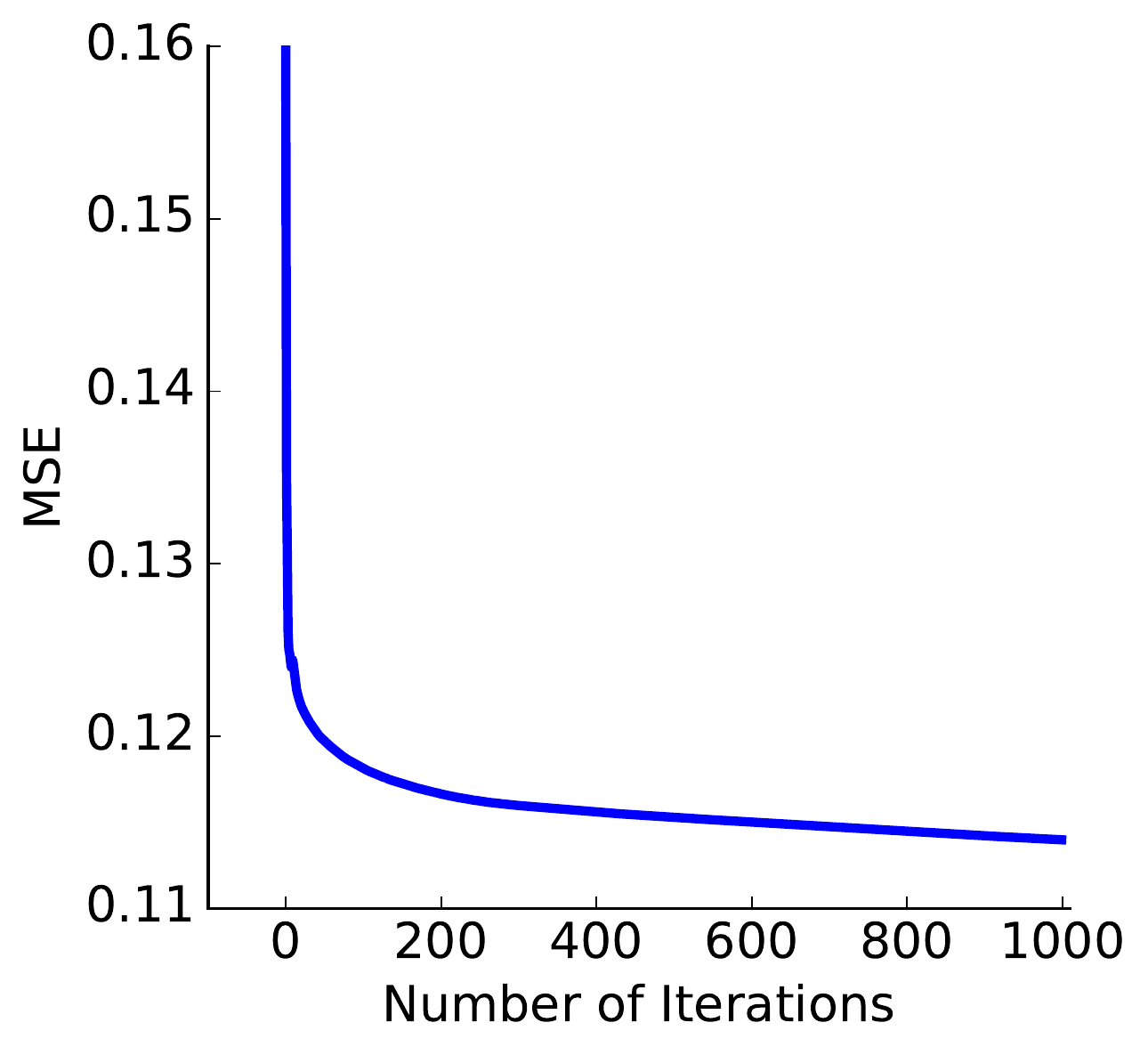}};
		\draw [anchor=north west] (.5\linewidth, .382\linewidth) node {\includegraphics[width=.469\linewidth]{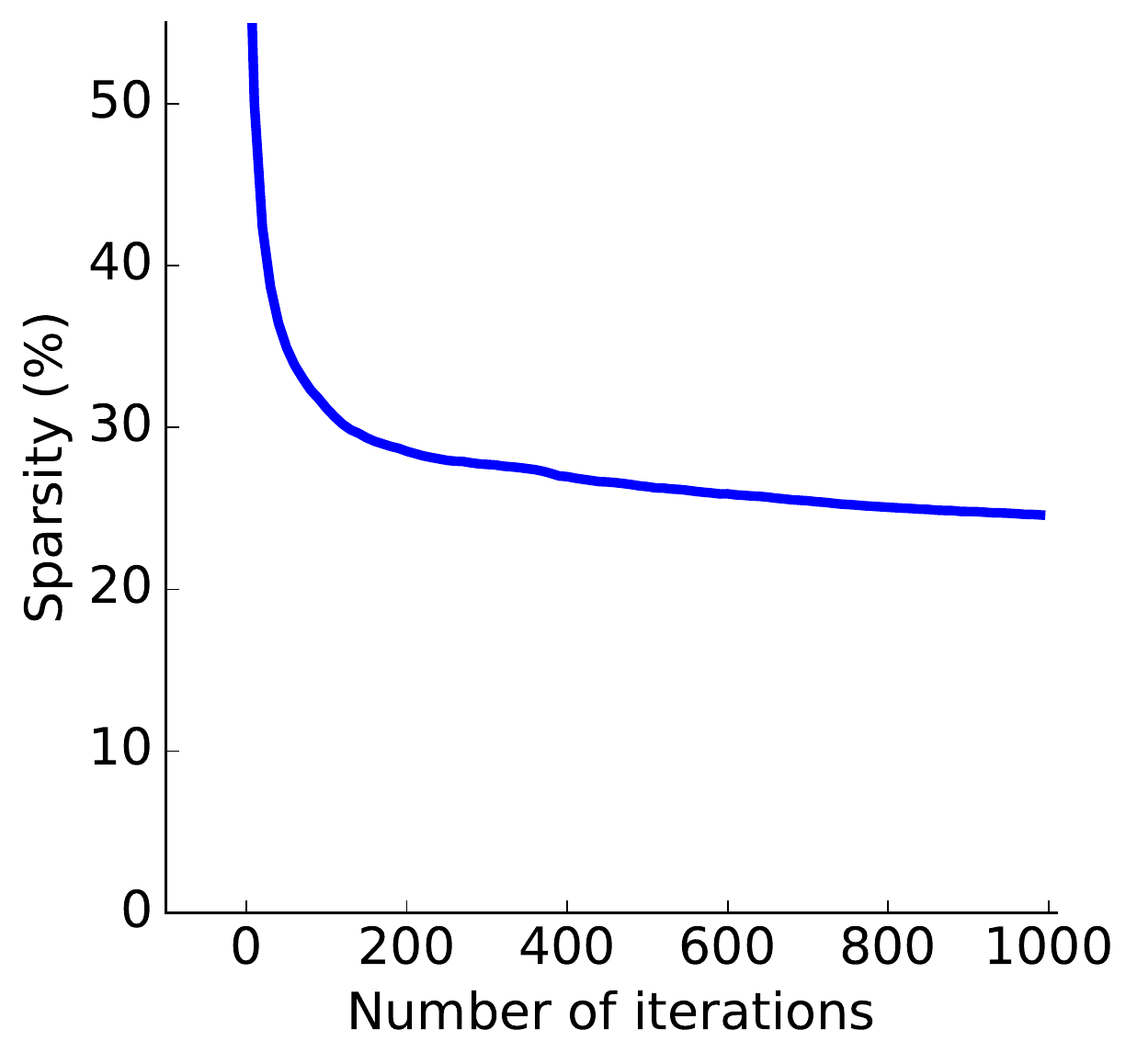}};
	\end{tikzpicture}    
	\caption
	{
	(a) Evolution of the mean square error between the reconstructed and the original database along the training process \textcolor{black}{(1000 trials)}.
	(b) Evolution of the sparsity (percentage of activated elements) of the second dictionary $D_{2}$.
	}
	\label{fig:fig1}
\end{figure}
Our model is composed of two convolutional layers: the first one contains $8$ filters of size $8\times 8$ pixels, and the second one contains $16$ filters of size $16\times16$ pixels. 
Consequently, the size of the deepest effective dictionary is $23\times 23$ pixels.
Our two-layered model was trained on $1000$ epochs. Figure~\ref{fig:fig1} shows the convergence of the representation error (Fig.~\ref{fig:fig1}a) and the evolution of the sparsity of the last layer dictionary during the training process (Fig.~\ref{fig:fig1}b). At the end of the training, the second dictionary $D_{2}$ has approximately 26\% of activated elements. 
Fig.~\ref{fig:fig2}b-c show the atoms of each dictionary at every layer. As observed in Fig.~\ref{fig:fig2}b, the \textcolor{black}{neuron's receptive fields of the first layer }
are oriented Gabor-like filters. This phenomenon was previously highlighted by~\cite{olshausen1996emergence} : orientation-selective filters comparable to the one found in the area V1 of the visual cortex~\cite{hubel1968receptive} tend to emerge from sparse coding strategies when applied on natural images. 
\textcolor{black}{Neuron's receptive fields of the second layer (shown in Fig.~\ref{fig:fig2}c) are made by the combination of the previous layer's receptive fields. As a consequence, neurons in the second layer are sensitive to more complex and specific stimuli.}
This increase in specificity and in the level of abstraction of the representation when going deeper in the network was also observed in the hierarchies of the cortical brain regions~\cite{mely2017towards}. \textcolor{black}{In particular, we found that these cells responded more to specific regions of the faces, as was also shown experimentally in human recordings~\cite{henriksson2015faciotopy}. }Fig.~\ref{fig:fig2}d shows reconstructed faces after the learning process. It can be seen that eyes, nose and contour of the face are well pronounced whereas textural features (e.g., hair, skin,~...) are smoothed. \textcolor{black}{Thus, the model is able to represent the input data without major loss of information with only a small number of activated element.}
\begin{figure}[!h]
\begin{tikzpicture}
	\centering
\draw [anchor=north west] (0.05\linewidth, 0.99\linewidth) node {\includegraphics[width=0.95\linewidth]{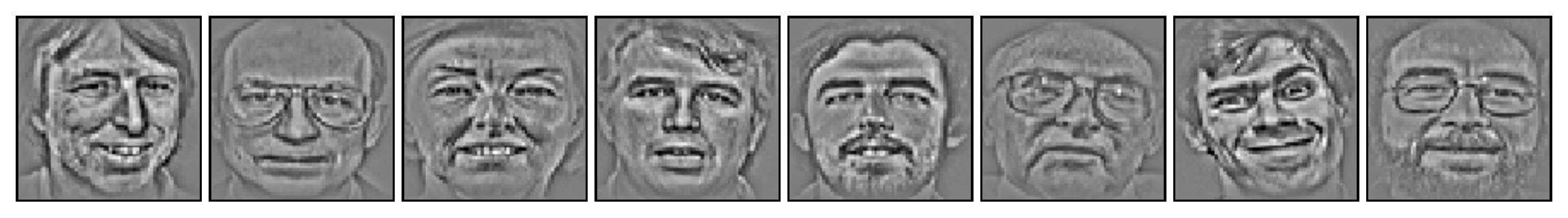}};
\draw [anchor=north west] (0.05\linewidth, .82\linewidth) node {\includegraphics[width=0.95\linewidth]{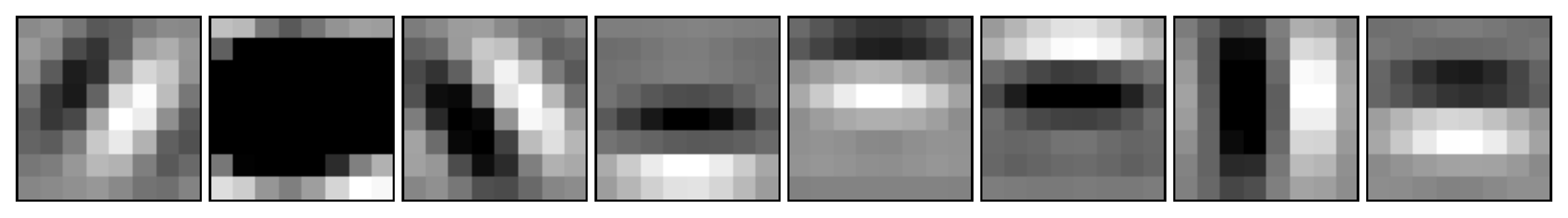}};
\draw [anchor=north west] (0.05\linewidth, .64\linewidth) node {\includegraphics[width=0.95\linewidth]{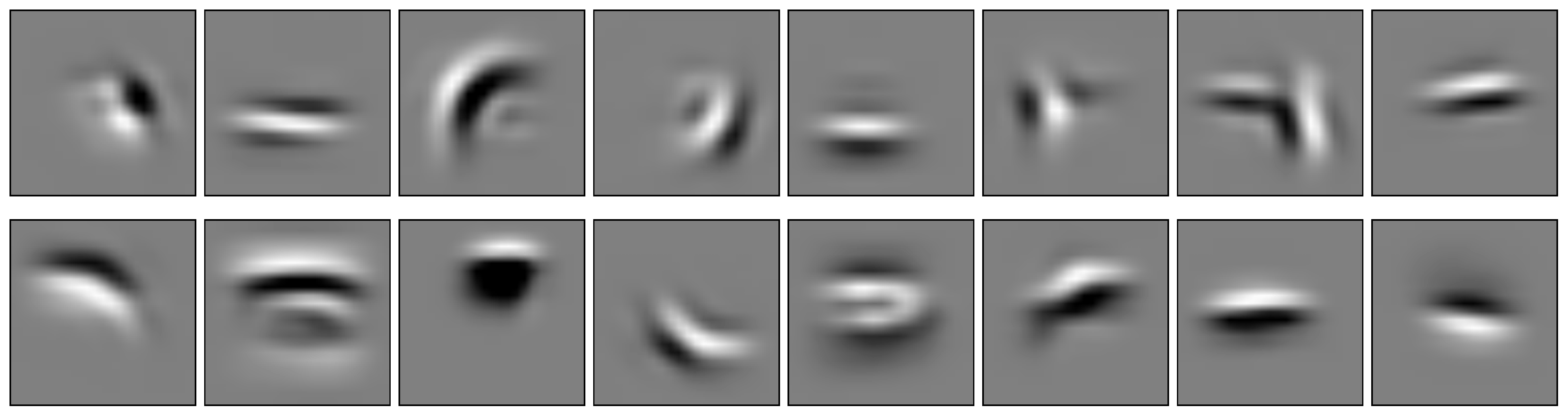}}; 
\draw [anchor=north west] (0.05\linewidth, .33\linewidth) node {\includegraphics[width=0.95\linewidth]{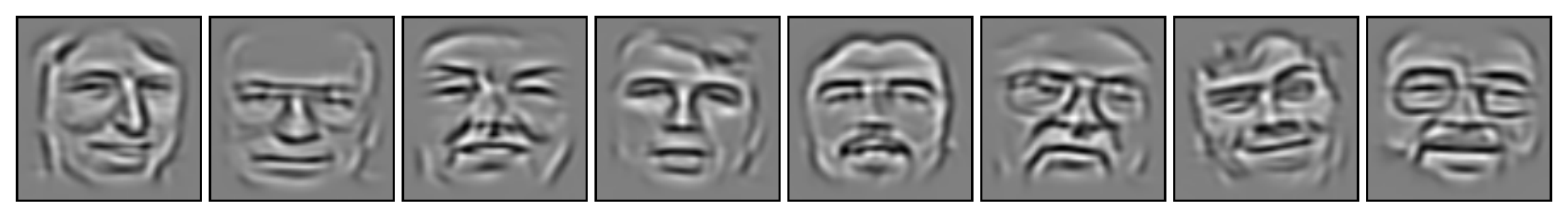}};
\begin{scope}
\draw [anchor=west,fill=white] (0.05\linewidth, 1\linewidth) node [above right=-3mm]{\small{(a) Examples of images after pre-processing}};
\draw [anchor=west,fill=white] (0.05\linewidth, .82\linewidth) node [above right=-3mm] {\small{(b) \textcolor{black}{Receptive fields of first layer's neurons}}}; 
\draw [anchor=west,fill=white] (0.05\linewidth, .65\linewidth) node [above right=-3mm] {\small{(c) \textcolor{black}{Receptive fields of second layer's neurons}}};
\draw [anchor=west,fill=white] (0.05\linewidth, .34\linewidth) node [above right=-3mm] {\small{(d) Example of Reconstructed Images}};
\end{scope}
\end{tikzpicture}
\caption
{
(a) \textcolor{black}{Pre-processed images from the AT$\&$T database. }
(b) The $8$ \textcolor{black}{receptive fields} (size $8\times 8$ pixels) composing the first layer dictionary ($D_{1}$) after the learning. They are similar to the Gabor-like receptive fields observed in the brain area V1.
(c) The $16$ \textcolor{black}{receptive fields} (size: $23\times 23$ pixels) of the second layer effective dictionary ($D^{(2)}$) at the end of the learning. 
(d) The reconstructed face once the network has learned the dictionary and the deeper sparse representation.
}
\label{fig:fig2}
\end{figure}
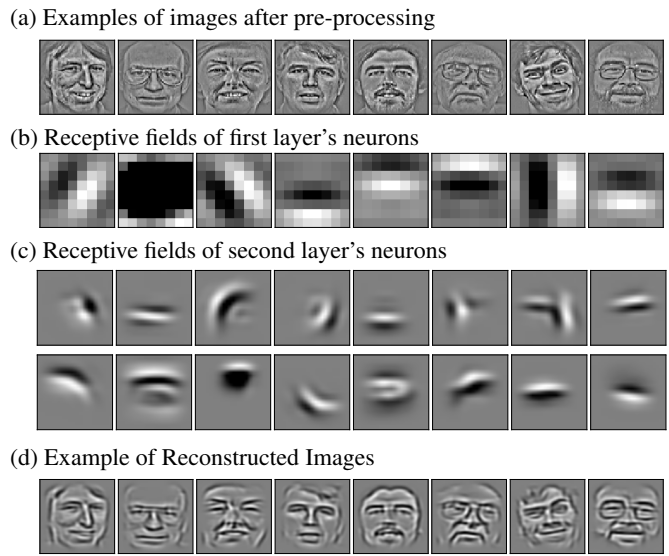
\vspace*{-.1cm}
\section*{Conclusion}
\label{sec:last-section}
\textcolor{black}{We have presented a hierarchical sparse coding algorithm to model the thalamo-cortical pathway. Interestingly, when trained on natural images, the model is optimally estimating the hierarchy of hidden physical causes (shapes, edges...) that constitute models of natural image generation. By increasing the scale and the specificity of receptive fields along the network, the model is able to combine simple and low level representation to build a more abstract and meaningful representation of the presented image. Further investigations concerning the classification performance of such a model need to be conducted to strengthen our comparison with the visual cortex.} 
%
\vspace*{-.25cm}
\section*{Acknowledgement and funding}
We thank M. Elad and J. Sulam for sharing the Matlab code of their ML-CDL implementation. {\Acknowledgments}
\newpage

\bibliographystyle{plainnat}
\bibliography{biblio}
\end{document}